\documentclass[conference]{IEEEtran}
\IEEEoverridecommandlockouts
\pdfoutput=1
\usepackage{cite}
\usepackage{amsmath,amssymb,amsfonts}
\usepackage{graphicx}
\usepackage{textcomp}
\usepackage{xcolor}
\usepackage{soul}
\usepackage{verbatim}
\usepackage{url}
\usepackage{color}
\usepackage{bm}
\usepackage[ruled]{algorithm}
\usepackage{amsthm}
\usepackage{float}
\usepackage{stfloats,algorithmicx,multirow,latexsym}
\usepackage{array}
\usepackage{mdwmath}
\usepackage{mdwtab}
\usepackage[font=footnotesize, caption=false]{subfig}

\newcounter{relctr} 
\everydisplay\expandafter{\the\everydisplay\setcounter{relctr}{0}} 
\newcommand\labelrel[2]{%
	\begingroup
	\refstepcounter{relctr}%
	\stackrel{\textnormal{(\alph{relctr})}}{\mathstrut{#1}}%
	\originallabel{#2}%
	\endgroup
}
\AtBeginDocument{\let\originallabel\label} 

\def\BibTeX{{\rm B\kern-.05em{\sc i\kern-.025em b}\kern-.08em
    T\kern-.1667em\lower.7ex\hbox{E}\kern-.125emX}}

\begin{document}

\title{Widely Linear Matched Filter: A Lynchpin towards the Interpretability of Complex-valued CNNs
}


\author{
	\IEEEauthorblockN{Qingchen Wang$^1$, Zhe Li$^1$, Zdenka Babic$^2$, Wei Deng$^3$, Ljubiša Stanković$^4$, Danilo P. Mandic$^5$}
	\IEEEauthorblockA{$^1$School of Electronic and Information Engineering, Soochow University, 215006 Suzhou, P. R. China}
	\IEEEauthorblockA{$^2$Department of Electrical Engineering, University of Banja Luka, 78000 Banja Luka, Bosnia and Herzegovina}
	\IEEEauthorblockA{$^3$College of Information Engineering, Jiaxing Nanhu University, 314001 Jiaxing, P. R. China}
	\IEEEauthorblockA{$^4$Department of Electrical Engineering, University of Montenegro, 81000 Podgorica, Montenegro}	
	\IEEEauthorblockA{$^5$Department of Electrical and Electronic Engineering, Imperial College London, SW7 2AZ London, U.K.}
	\IEEEauthorblockA{qcwangmail@stu.suda.edu.cn, lizhe@suda.edu.cn, zdenka.babic@etf.unibl.org,\\ wei.deng@jxnhu.edu.cn,ljubisa@ac.me, d.mandic@imperial.ac.uk}
}

\maketitle

\begin{abstract}
A recent study on the interpretability of real-valued convolutional neural networks (CNNs) \cite{Stankovic_Mandic_2023CNN} has revealed a direct and physically meaningful link with the task of finding features in data through matched filters. However, applying this paradigm to illuminate the interpretability of complex-valued CNNs meets a formidable obstacle: the extension of matched filtering to a general class of noncircular complex-valued data, referred to here as the widely linear matched filter (WLMF), has been only implicit in the literature. To this end, to establish the interpretability of the operation of complex-valued CNNs, we introduce a general WLMF paradigm, provide its solution and undertake analysis of its performance. For rigor, our WLMF solution is derived without imposing any assumption on the probability density of noise. The theoretical advantages of the WLMF over its standard strictly linear counterpart (SLMF) are provided in terms of their output signal-to-noise-ratios (SNRs), with WLMF consistently exhibiting enhanced SNR. Moreover, the lower bound on the SNR gain of WLMF is derived, together with condition to attain this bound. This serves to revisit the convolution-activation-pooling chain in complex-valued CNNs through the lens of matched filtering, which reveals the potential of WLMFs to provide physical interpretability and enhance explainability of general complex-valued CNNs. Simulations demonstrate the agreement between the theoretical and numerical results. 
\end{abstract}

\vspace{-1mm}
\section{Introduction}\label{sec:intro}
\vspace{-1mm}
Complex-valued neural networks (NNs) offer an accurate representation of both magnitude and phase features from data \cite{mandic2009}. This leads to superior efficiency compared to their real-valued counterparts in applications such as speech enhancement, interferometric synthetic aperture radar image interpretation, and wind prediction \cite{hirose2012complex,chiheb2017deep,Kang2020,tachibana2018wind}. Despite great successes of deep artificial NNs in the last decade, the interpretability issue which arises from their black box nature remains a bottleneck which prevents a more widespread adoption of NNs \cite{fan2021interpretability}. A recent study connects the well-understood (and theoretically supported) concept of matched filter with feature identification mechanism in real-valued convolutional neural networks (CNNs). This provides an innovative and physically meaningful interpretation of the building blocks in real-valued CNNs, namely the convolution-activation-pooling chain and corresponding learning strategies \cite{Stankovic_Mandic_2023CNN,Stankovic_Mandic_2023GCNN}. This recent insight naturally prompts us to investigate the potential of the matched filtering paradigm in demystifying complex-valued CNNs---a subject of this work.

The matched filter (MF) is a fundamental detection technique in statistical signal processing and machine learning \cite{Turin1960,Turin1976}. While the term MF has been used across diverse domains over the decades, its exact meaning varies depending on the specific context. Indeed, alternative names such as \textit{template matching} in medical imaging \cite{brunelli2009template,ahn1999adaptive,Arias2022}, \textit{maximum ratio combining} in communications \cite{joham2005,Yang_Hanzo_2015,Hama2019}, and \textit{matched subspace detector} in target detection \cite{Scharf1994,Kammoun2018,Ghasemi2020,Neinavaie_Khalife2022}, to name but a few, illustrate the diverse interpretations of the underlying concept. Definitions of the MF in these studies have been developed from two different points of view. The first, rooted in the principle of the generalized likelihood ratio test in signal processing and machine learning, views the MF as a sufficient statistic for choosing between two hypothesis: ${\mathcal{H}}_0$, representing noise only, and ${\mathcal{H}}_1$, representing signal presence within noise \cite[Sec. 4.9]{Scharf_SSP_book}. In this sense, the MF is demonstrated to be optimal under the Gaussian background. The second perspective derives the MF as a cross-correlator, which maximizes the signal-to-noise-ratio (SNR), also referred to as \textit{deflection} in some literature \cite{Picinbono_1995_TAES,Schreier_2005_TIT}, thus addressing scenarios where the probability densities of the observation under ${\mathcal{H}}_0$ or ${\mathcal{H}}_1$ are not completely known \cite[Sec. 4.2]{richards2014fundamentals}. 

The basic MF problem based on the SNR maximization criterion is stated as follows. 
Let $r(n)$, $n \in \{1,\dots,N\}$, denote a complex-valued observed signal of length $N$, described as 
\begin{align}\label{sl:r1}
	r(n)=x(n)+v(n),
\end{align}
where the  desired input $x(n)$ is a known deterministic signal\footnote{The MF can process either a deterministic or a random input with known statistics (the latter is referred to as stochastic MF \cite{cavassilas1991}). Without loss in generality, we limit our scope to matched filtering for deterministic signals.}, the additive noise $v(n)$ is a zero-mean random variable, and both are complex-valued. Assume that the observed signal, $r(n)$, is filtered by an $L$-tap FIR filter, of which the coefficients, ${\bf{f}}\in {\mathbb{C}}^{L\times 1}$, $L \le N$, are uncorrelated with noise $v(n)$. Then, at a sample index $n_p$, $n_p \in \{L,\dots,N\}$, the filter output $y(n_p)$ can be written as
\begin{align}\label{eq:yn_p_SLMF}
	y(n_p)&={\bf{f}}^H {\bf{r}}={\bf{f}}^H{\bf{x}}+{\bf{f}}^H{\bf{v}},
\end{align}
where ${\bf{r}} \triangleq [r(n_p),r(n_p-1),\dots,r(n_p-L+1)]^T \in \mathbb{C}^{L\times1}$ is the observed measurement vector at $n_p$, and ${\bf{x}} \triangleq [x(n_p),x(n_p-1),\dots,x(n_p-L+2),x(n_p-L+1)]^T \in \mathbb{C}^{L\times1}$ is the input vector, with ${\bf{v}} \triangleq [v(n_p),v(n_p-1),\dots,v(n_p-L+1)]^T \in \mathbb{C}^{L\times1}$ as the noise vector. The goal is to design a filter of which the output has an evident peak at $n=n_p$ within the observation window $0\le n<N$, thereby confirming the existence and location of the deterministic signal $\bf{x}$. This is achieved by maximizing the output SNR at a sample index $n_p$, defined as \cite[Sec. 5.4.1]{therrien1992discrete}, 
%
\begin{equation}\label{eq:SLMFprob}
	\text{SNR}_{\text{SL}}(n_p) \triangleq \frac{|{\bf{f}}^H{\bf{x}}|^2}{E\left[|{\bf{f}}^H{\bf{v}}|^2\right]},
\end{equation}
where $E[\cdot]$ is the statistical expectation operator. For conciseness, the sample index $n_p$ is omitted in the sequel. The solution to this SNR maximization problem is the set of MF coefficients \cite[Eq. (5.56)]{therrien1992discrete}
\begin{align}\label{eq:slmf} 
	\hat {\bf{f}}=\alpha {\bf{R}}_{v}^{-1}{\bf{x}},
\end{align}
where $\alpha$ is a real-valued scaling factor
and ${\bf{R}}_v \triangleq E[{\bf{v}}{\bf{v}}^H] \in {\mathbb{C}}^{L\times L}$ the covariance matrix of the noise vector ${\bf v}$. Upon plugging \eqref{eq:slmf} into \eqref{eq:SLMFprob}, the maximum output SNR is obtained as \cite[Eq. (5.55)]{therrien1992discrete}
\begin{align}\label{eq:SNRSLMF}
	\text{SNR}_{\text{SLMF}}={\bf{x}}^H{\bf{R}}_v^{-1}{\bf{x}}. 
\end{align}
%

The so derived MF, referred to as the strictly linear MF (SLMF), is built upon an implicit assumption that the noise, $v(n)$, is a second-order circular (proper) signal\footnote{A zero-mean complex-valued random variable, $a$, is called proper if its complementary variance $E[aa]=0$, otherwise it is called improper.}. While this assumption is justified for many scenarios in signal processing and machine learning, it is admittedly restrictive in a broader context. A typical example of noise noncircularity exists in radar and wireless communications, where the inherently second order noncircular (improper) interference from other sources is treated as noise \cite{Chevalier2006,Li2021,Lameiro2017,Yu2020_TC}. Therefore, for general improper complex-valued random processes, both the covariance matrix, $\bold{{R}}=E[{\bold{{x}}}{\bold{{x}}}^H]$, and the pseudo-covariance matrix, ${\bold{{P}}}=E[{\bold{{x}}}{\bold{{x}}}^T]$, should be considered to completely exploit the full second-order statistics within the complex domain $\mathbb{C}$. In practice, this is achieved by widely linear modeling \cite{mandic2009,schreier2010statistical}. The advantage of widely linear filtering over its conventional strictly linear counterpart was initially demonstrated in \cite{Picinbono1995}, where explicit expressions for the improvement in mean square estimation (MSE) are derived. This has triggered a large volume of research on noncircular complex data and improper systems, ranging adaptive filtering \cite{Chevalier_2021,zhangxu2023} to source separation \cite{Lee2022,Meziani2023}, through to wireless transmission \cite{Yu2020_TC,Liu_IGS_2023}.

However, extensions of SLMF to the processing of general improper complex-valued data have been largely lacking. Indeed, the widely linear MF (WLMF) was only implicitly assumed in the derivation of widely linear equalization schemes for frequency-selective channels \cite{Gerstacker2003} or when dealing with noncircular interference in the single antenna interference cancellation problem \cite{Chevalier2006}. Notably, the work in \cite{Schreier_2005_TIT} investigates the detection and estimation of improper complex random signals. It demonstrates that, compared to strictly linear processing, the WL framework can achieve a performance gain of up to a factor of two higher in the presence of additive white Gaussian noise.

Recognizing the pivotal role WLMF plays in demystifying complex-valued CNNs, the present work aims to fill the gap related to a rigorous derivation and analysis of WLMF and, in particular, the quantification of its performance advantage over the SLMF. To this end, we first extend the basic MF problem setting to accommodate the generality of complex-valued signals and derive the expression of the WLMF based on the SNR maximization criterion. This is achieved without imposing any specific assumptions on the probability density of the noise. Next, by analyzing the SNR gain of the WLMF over SLMF, we show that the WLMF always yields an SNR improvement, except for a null input. Moreover, our analysis not only demonstrates that the SNR of SLMF is exactly half of its WLMF counterpart in the case of proper noise, but also establishes a lower bound on the SNR gain, which can be arbitrarily large for highly improper noise. 
Finally, we introduce the  interpretation of the convolution-activation-pooling chain in complex-valued CNNs through the lens of matched filtering. Simulations on a complex-valued CNN performing an identification task demonstrate that both the theoretical and practical performance advantages of WLMF over SLMF translate to enhanced interpretability and classification capability of CNNs. 

The rest of the paper is organized as follows. Section II presents the derivation of the WLMF. In Section III, the analysis of the WLMF is performed through a comparison of its SNR with that of the SLMF. Numerical examples are given in Section IV to validate the operation of WLMF. Interpretation of complex-valued CNNs from the perspective of WLMF are provided in Section V. Finally, Section VI concludes the paper.

\vspace{-1mm}
\section{Widely linear matched filter}\label{sec:format}
\vspace{-1mm}
%
 \begin{figure}[!t]
	 	\centering
	 	\includegraphics[width=0.4\textwidth]{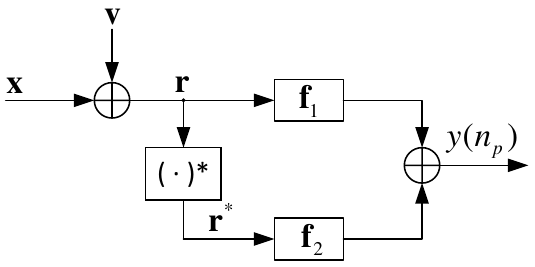}
	 	\caption{Block diagram of the widely linear matched filter.}
	 	\label{fig:1}
	 \end{figure}
Consider a WLMF model, whereby the input-output relationship remains the same as in \eqref{sl:r1}, apart from that the noise $v(n)$ is allowed to be improper. The block diagram of the considered WLMF model is depicted in Fig. \ref{fig:1}, where a pair of $L$-tap WLMFs, with weight vectors ${\bf{f}}_1$ and ${\bf{f}}_2$, are applied to jointly process the observed signal $r(n)$ and its complex conjugate $r^*(n)$. Consequently, the filter output at a sample index $n_p$ can be expressed as
\begin{align}\label{y:2}
	y(n_p) = {\bf{f}}_1^H{\bf{r}}+{\bf{f}}_2^H{\bf{r}}^*.
\end{align}
Following the WL model \cite[Sec. 13.1.1]{mandic2009}, upon inserting \eqref{sl:r1} into \eqref{y:2}, we obtain
\begin{equation}\label{eq:yn3}	
	y(n_p)={\bf{f}}_1^H{\bf{x}}+{\bf{f}}_2^H{\bf{x}}^*+{\bf{f}}_1^H{\bf{v}}+{\bf{f}}_2^H{\bf{v}}^*.
\end{equation}
For compactness, the augmented WLMF weight vector, ${\bf {w}} \in {\mathbb{C}}^{2L\times 1}$, the augmented input vector, ${\bf{z}}\in {\mathbb{C}}^{2L\times 1}$, and the augmented noise vector, ${\bf{q}}\in {\mathbb{C}}^{2L\times 1}$, are defined as
\begin{equation}\label{eq:augdef}
	{\bf {w}}\triangleq[{\bf{f}}_1^T,{\bf{f}}_2^T]^T,\quad {\bf{z}}\triangleq[{\bf{x}}^T,{\bf{x}}^H]^T, \quad{\bf{q}}\triangleq[{\bf{v}}^T,{\bf{v}}^H]^T,   
\end{equation}
so that the widely linear model in \eqref{eq:yn3} becomes
\begin{equation}\label{eq:ynaug}
	y(n_p)={\bf {w}}^H{\bf{z}}+{\bf {w}}^H{\bf{q}}.
\end{equation}
Following the definition in \eqref{eq:SLMFprob}, the output SNR of the WLMF can be obtained from \eqref{eq:ynaug} as
\begin{align}\label{y:5}
	{\text{SNR}}_{\text{WL}}=\frac{|{\bf {w}}^H{\bf{z}}|^2}{E\big[|{\bf {w}}^H{\bf{q}}|^2\big]}.
\end{align}
The problem in \eqref{y:5} now becomes that of the maximization of ${\text{SNR}}_{\text{WL}}$ by choosing the appropriate vector ${\bf {w}}$, whereby the denominator $E\big[|{\bf {w}}^H{\bf{q}}|^2\big]$ can be expanded as
\begin{align}\label{y:6}
	E\left\{|{\bf {w}}^H{\bf{q}}|^2\right\}={\bf {w}}^H{\bf{R}}_{q}{\bf {w}},
\end{align}
where ${\bf{R}}_{q} \in \mathbb{C}^{2L\times 2L}$ is the augmented noise covariance matrix, defined as
\begin{equation}\label{R:1}
	\begin{aligned}
		{\bf{R}}_{q}&\triangleq E[{\bf{q}}{\bf{q}}^H]=\begin{bmatrix}
			{\bf{R}}_v&{\bf{C}}_v \\
			{\bf{C}}_v^*& {\bf{R}}_v^*
		\end{bmatrix}.
	\end{aligned}
\end{equation}
Observe that the matrix ${\bf{R}}_{q}$ contains the noise covariance matrix, ${\bf{R}}_v$, as well as the noise complementary covariance matrix, ${\bf{C}}_v \triangleq E\left\{{\bf{v}}{\bf{v}}^T\right\} \in {\mathbb{C}}^{L\times L} $. A substitution of \eqref{y:6} into \eqref{y:5} further gives
\begin{align}\label{snr:23}
	{\text{SNR}}_{\text{WL}} &=\frac{|{\bf {w}}^H{\bf{z}}|^2}{{\bf {w}}^H{\bf{R}}_{q}{\bf {w}}}\nonumber\\
	&=\frac{|({\bf{R}}^{1/2}_{q}{\bf {w}})^H({\bf{R}}^{-1/2}_{q}{\bf{z}})|^2}{({\bf{R}}^{1/2}_{q}{\bf {w}})^H({\bf{R}}^{1/2}_{q}{\bf {w}})}\nonumber\\
	&\labelrel \leq{eqrel:2}  \frac{({\bf{R}}^{1/2}_{q}{\bf {w}})^H({\bf{R}}^{1/2}_{q}{\bf {w}})\big({\bf{R}}^{-1/2}_{q}{{\bf z}})^H({\bf{R}}^{-1/2}_{q}{\bf z}\big)}{({\bf{R}}^{1/2}_{q}{\bf {w}})^H({\bf{R}}^{1/2}_{q}{\bf {w}})},
\end{align}
where \eqref{eqrel:2} follows from the Cauchy-Schwartz inequality. The equality sign holds if and only if ${\bf{R}}^{1/2}_{q}{\bf {w}}=\beta {\bf{R}}^{-1/2}_{q}{\bf{z}}$,
where $\beta$ is a real-valued scaling factor. Therefore, the augmented weight vector of the WLMF is obtained as
\begin{align}\label{y:8}
	\hat{\bf {w}}=\beta {\bf{R}}_{q}^{-1}{\bf{z}}.
\end{align}
Upon plugging \eqref{y:8} into \eqref{y:5}, the maximum SNR of the WLMF is obtained as
\begin{equation}\label{eq:SNRWLMF}
	\text{SNR}_{\text{WLMF}}={\bf z}^H{\bf{R}}^{-1}_{q}{\bf z}.
\end{equation}
Through matrix inversion, the inverse of the matrix ${\bf{R}}_{q}^{-1}$ can be partitioned into a block matrix as \cite[Sec. A1.3.1]{schreier2010statistical}
\begin{normalsize}
	\begin{align}\label{eq:Rbinv}
		\setlength{\arraycolsep}{0.1pt}
		\begin{array}{lc}
			{\bf{R}}^{-1}_{q}=&\\
			\begin{bmatrix}\!
				\!({\bf{R}}_v\!-\!{\bf{C}}_v{\bf{R}}_v^{-*}{\bf{C}}_v^*)^{-1}\!&\!\!\!\!\!\!\!\!-({\bf{R}}_v\!-\!{\bf{C}}_v{\bf{R}}_v^{-*}{\bf{C}}_v^*)^{-1}{\bf{C}}_v{\bf{R}}_v^{-*} &\\
				-({\bf{R}}_v^*\!-\!{\bf{C}}_v^*{\bf{R}}_v^{-1}{\bf{C}}_v)^{-1}{\bf{C}}_v^*{\bf{R}}_v^{-1}\!&\! ({\bf{R}}_v^*\!-\!{\bf{C}}_v^*{\bf{R}}_v^{-1}{\bf{C}}_v)^{-1}
			\end{bmatrix}.
		\end{array}
	\end{align}
\end{normalsize}
Finally, a substitution of \eqref{eq:Rbinv} and \eqref{eq:augdef} into \eqref{y:8} yields the pair of WLMF weight vectors
\begin{align}
	{\hat{\bf{f}}}_1&=\beta({\bf{R}}_v-{\bf{C}}_v{\bf{R}}_v^{-*}{\bf{C}}_v^*)^{-1}[{\bf{x}}-{\bf C}_v{\bf{R}}_v^{-*}{\bf{x}}^*],\label{eq:f1}\\
	{\hat {\bf{f}}}_2&=\beta({\bf{R}}_v^*-{\bf{C}}_v^*{\bf{R}}_v^{-1}{\bf{C}}_v)^{-1}[{\bf{x}}^*-{\bf{C}}_v^*{\bf{R}}_v^{-1}{\bf{x}}]\label{eq:f2}.
\end{align}

{\textit{Remark 1}}: An inspection of \eqref{eq:f1} and \eqref{eq:f2} shows that the pair of WLMF weight vectors are mutually conjugate, i.e., ${\bf{f}}_1={\bf{f}}_2^*$. In a particular case when the noise $v(n)$ is proper, that is, ${\bf{C}}_v = {\bf 0}$, expression \eqref{eq:f1} reduces to 
\begin{align}
	{\hat {\bf{f}}}_1&=\beta{\bf{R}}_v^{-1}{\bf{x}},
\end{align}
which boils down into the expression for the standard SLMF weight vector $\hat {\bf{f}}$ in \eqref{eq:slmf}, thus confirming the generality of our result.
\vspace{-1mm}
\section{Performance analysis of the WLMF}
\label{sec:ana}
\vspace{-1mm}
For rigor, we now proceed to quantify the SNR gain of WLMF over the SLMF and establish the lower bound of the SNR gain advantage. 

\textit{Theorem 1}: The SNR of the WLMF is always greater than that of the SLMF, unless the input vector ${\bf{x}}$ is null.

\textit{Proof:} Define the SNR gain of WLMF over SLMF as
\begin{equation}\label{eq:SNRGainDef}
	\text{SNR}_{\Delta} \triangleq \text{SNR}_{\text{WLMF}}-\text{SNR}_{\text{SLMF}}.
\end{equation}
Upon inserting \eqref{eq:SNRWLMF} and \eqref{eq:SNRSLMF} into \eqref{eq:SNRGainDef}, we obtain
\begin{align}\label{eq:delta_snr}
	\text{SNR}_{\Delta}=&{\bf{z}}^H{\bf{R}}_q^{-1}{\bf{z}}
	-{\bf{x}}^H{\bf{R}}_v^{-1}{\bf{x}}\nonumber\\
	=&{\bf{f}}_1^H{\bf{x}}(n)+{\bf{f}_2}^H{\bf{x}}^*(n)-{\bf{x}}^H{\bf{R}}_v^{-1}{\bf{x}}\nonumber\\
	=&\big({\bf{x}}^*-{\bf{C}}_v^*{\bf{R}}_v^{-1}{\bf{x}}\big)^H({\bf{R}}_v^*-{\bf{C}}_v^*{\bf{R}}_v^{-1}{\bf{C}}_v)^{-1}\nonumber\\&\times \big({\bf{x}}^*-{\bf{C}}_v^*{\bf{R}}_v^{-1}{\bf{x}}\big).
\end{align}
Observe that the matrix ${\bf{R}}_v^*-{\bf{C}}_v^*{\bf{R}}_v^{-1}{\bf{C}}_v$ is positive definite since it is the Schur complement of the noise covariance matrix ${\bf{R}}_v$. Therefore, the SNR gain $\text{SNR}_{\Delta}$ is always positive unless the input vector ${\bf x} = {\bf 0}$. \hfill\qedsymbol

{\textit{Corollary 1}:} For a proper noise, $v(n)$, the SNR of the WLMF is twice that of its SLMF counterpart.

{\textit{Proof:}} Proper noise $v(n)$ has a vanishing complementary statistics, i.e., ${\bf C}_v={\bf 0}$, hence the SNR gain in \eqref{eq:delta_snr} becomes
 \begin{align}\label{eq:delta_snr_circ}
	 	\text{SNR}_{\Delta}={\bf{x}}^T{\bf{R}}_v^{-*}{\bf{x}}^*.
 \end{align}
Since the SNR is inherently real-valued, we obtain $\text{SNR}_{\Delta} = \text{SNR}_{\text{SLMF}} = \frac{1}{2}\text{SNR}_{\text{WLMF}}$. \hfill\qedsymbol

\textit{Remark 2}: Note that the SNR gain in \eqref{eq:delta_snr} admits a similar form to the MSE improvement given in \cite[Eq. (2.10)]{Picinbono1995}, despite the different inherent physical mechanisms. The SNR advantage in the case of proper noise, as established in \textit{Corollary 1}, is consistent with the analysis in \cite{Schreier_2005_TIT}, which indicates that widely linear processing attains a 3 dB gain over SL processing in detecting a coherent signal -- a category to which the considered deterministic signal $x(n)$ belongs -- over a non-coherent signal. Notably, our analysis attains this gain without imposing any assumption on the probability density of the noise. 
%
%


We next provide further insights into the expression \eqref{eq:delta_snr} so as to establish a lower bound on the SNR gain $\text{SNR}_{\Delta}$. To this end, we shall first introduce some necessary definitions. The approximate uncorrelating (AUT) transform \cite{CheongTook2012} states that the unitary matrix ${\bf Q }\in {\mathbb{C}}^{L\times L}$, which is used in the Takagi/Autonne factorization \cite[Ch. 3]{Hornbook1991} to diagonalize the complex-valued symmetric complementary covariance matrix ${\bf{C}}_v$, can also be used to approximately diagonalize the covariance matrix ${\bf{R}}_v$, to give
%
\begin{subequations}\label{eq:AUT}
	\begin{align}
	{\bf{C}}_v &=  {\bf Q}  {\bm \Lambda}_c {\bf Q}^T,\\
	{\bf{R}}_v &\simeq  {\bf Q}  {\bm \Lambda}_r {\bf Q}^H,
	\end{align}
\end{subequations}
where ${\bm \Lambda}_c = {\text{diag}} \{p_1, p_2, \dots, p_L\}$ is a diagonal matrix of real-valued entries, $p_1 \ge p_2 \ge \dots \ge p_L$ which are the nonnegative square roots of ${\bf{C}}_v{\bf{C}}_v^H$, with ${\bm \Lambda}_r = {\text{diag}} \{ \lambda_1, \lambda_2, \dots, \lambda_L\}$, and $\lambda_1 \ge\lambda_r\ge \dots \ge \lambda_L$ as the real-valued eigenvalues of ${\bf{R}}_v$. By the AUT, the rotated input vector can be defined as
\begin{equation}\label{eq:tildebfx}
	{\tilde {\bf x}} \triangleq {\bf Q}^H {\bf x} = [{\tilde x}_1,{\tilde x}_2,\dots,{\tilde x}_L]^T.    
\end{equation}
We now introduce a vector ${\bm \epsilon} \triangleq [\epsilon_1,\epsilon_2,\dots,\epsilon_L]^T$, as 
\begin{equation}\label{eq:epsilon}
	\epsilon_i \triangleq \frac{\Re\{{\tilde x}_i{\tilde x}_i\}}{{\tilde x}_i{\tilde x}_i^*},
\end{equation}
where the elements $\epsilon_i$ are termed the \textit{power difference coefficients}. These represent the normalized power difference between the real and imaginary channels of the $i$th rotated input ${\tilde x}_i$, while the operator $\Re\{\cdot\}$ in \eqref{eq:epsilon} extracts the real part of a complex variable. Observe that $-1 \le \epsilon_i \le 1$, and $\epsilon_i = 0$ holds only when ${\tilde x}_i$ has equal powers in its real and imaginary channels (properness). We also define a vector ${\bm \rho} \triangleq [\rho_1,\rho_2,\dots,\rho_L]^T$, with elements
\begin{equation}\label{eq:rho}
	\rho_i \triangleq \frac{p_i}{\lambda_i},    
\end{equation}
which represents the degree of impropriety of the $i$th noise component \cite{Ollila2008,Mandic2015}. Note that $\rho_i$ satisfies $0\le \rho_i \le 1$, with the bounds $0$ and $1$ indicating respectively the proper and maximally improper cases  \cite[Sec. 1.6.2]{schreier2010statistical}.

\textit{Theorem 2}: The SNR gain approximated by the AUT, $\text{SNR}_{\Delta}^{\text{ap}}$, achieves its lower bound if and only if $\forall i \in \{1,2,\dots, L\}$ the degree of impropriety of noise, $\rho_i$, satisfies
\begin{equation}\label{eq:cond}
	\rho_i = \left\{ \begin{matrix}
		0 & (-1 \le \epsilon_i\le 0)  \\
		\frac{1-\sqrt{1-\epsilon_i^2}}{\epsilon_i} & (0<\epsilon_i\le 1). 
	\end{matrix} \right.
\end{equation}

\textit{Proof:} Upon inserting \eqref{eq:AUT} into \eqref{eq:delta_snr}, we obtain 
\begin{align}\label{eq:SNRdelta1}
	\text{SNR}_{\Delta}^{\text{ap}} = & ({\bf x}^*-{\bf Q}^*{\bm \Lambda}_c{\bf Q}^H{\bf Q}^{-H}{\bm \Lambda}_r^{-1}{\bf Q}^{-1}{\bf x})^H\nonumber\\
	& \times \big({\bf Q}^*{\bm \Lambda}_r{\bf Q}^T\nonumber\\
	&-{\bf Q}^*{\bm \Lambda}_c{\bf Q}^H{\bf Q}^{-H}{\bm \Lambda}_r^{-1}{\bf Q}^{-1}{\bf Q}{\bm \Lambda}_c{\bf Q}^T\big)^{-1} \nonumber\\
	&\times ({\bf x}^*-{\bf Q}^*{\bm \Lambda}_c{\bf Q}^H{\bf Q}^{-H}{\bm \Lambda}_r^{-1}{\bf Q}^{-1}{\bf x})\nonumber\\
	=& ({\bf x}^*-{\bf Q}^*{\bm \Lambda}_c{\bm \Lambda}_r^{-1}{\bf Q}^{H}{\bf x})^H\nonumber\\
	&\times \big({\bf Q}^*({\bm \Lambda}_r^2-{\bm \Lambda}_c^2){\bm \Lambda}_r^{-1}{\bf Q}^T\big)^{-1}\nonumber\\
	&\times ({\bf x}^*-{\bf Q}^*{\bm \Lambda}_c{\bm \Lambda}_r^{-1}{\bf Q}^{H}{\bf x})\nonumber\\
	=&({\tilde {\bf x}}^*-\frac{{\bm \Lambda}_c}{{\bm \Lambda}_r}{\tilde {\bf x}})^H\frac{{\bm \Lambda}_r}{{\bm \Lambda}_r^2-{\bm \Lambda}_c^2}({\tilde {\bf x}}^*-\frac{{\bm \Lambda}_c}{{\bm \Lambda}_r}{\tilde {\bf x}}).
\end{align}

The expansion of the quadratic form in \eqref{eq:SNRdelta1}  gives
\begin{align}\label{eq:SNRdelta2}
	\text{SNR}_{\Delta}^{\text{ap}} 
	\!=\! \sum\limits_{i=1}^{L} \frac{{\lambda_i}}{{\lambda_i^2 - p_i^2}} \!\left[|{\tilde x}_i|^2\left(1\!+\!\frac{p_i^2}{\lambda_i^2}\right)\!-\!2\frac{p_i}{\lambda_i}\Re\{{\tilde x}_i^2\}\right].
\end{align}
A substitution of the \textit{power difference coefficient} in \eqref{eq:epsilon} and the degree of impropriety in \eqref{eq:rho} into \eqref{eq:SNRdelta2} yields
\begin{equation}\label{eq:SNRdelta3}
	\text{SNR}_{\Delta}^{\text{ap}} = \sum\limits_{i=1}^{L} \frac{|{\tilde x}_i|^2}{\lambda_i} g(\rho_i),
\end{equation}
where
\begin{equation}\label{eq:grhoi}
	g(\rho_i) \triangleq  \frac{1+\rho_i^2-2\epsilon_i\rho_i}{1-\rho_i^2}.   
\end{equation}
Observe from \eqref{eq:SNRdelta3} that the impact of the noise impropriety on the SNR gain rests upon the function $g(\rho_i)$. To this end, we shall investigate the first derivative of $g(\rho_i)$ with respect to $\rho_i$. After some algebraic manipulations, we arrive at
%
\begin{equation}\label{eq:1orderderiv}
	\frac{d g(\rho_i)}{d \rho_i} = -2\frac{\epsilon_i\rho_i^2-2\rho_i+\epsilon_i}{(1-\rho_i^2)^2}.    
\end{equation}
Since the \textit{power difference coefficient}, $\epsilon_i$, plays a role in the monotonicity of the function, $g(\rho_i)$, we next split up the analysis into two cases.

\textit{Case 1):} $-1<\epsilon_i\le 0$. In this case, the numerator on the RHS of \eqref{eq:1orderderiv} satisfies $\epsilon_i\rho_i^2-2\rho_i+\epsilon_i \le 0$, therefore $\frac{d g(\rho_i)}{d \rho_i} \ge 0$. Observe that the equality sign holds if and only if
\begin{equation}
	\rho_i=0 \quad \text{and} \quad \epsilon_i=0.   
\end{equation}
This shows that $g(\rho_i)$ is a monotonically increasing function of $\rho_i$, with the minimum achieved when $\rho_i=0$. 

\textit{Case 2):}  $0<\epsilon_i\le 1$. Let $\frac{d g(\rho_i)}{d \rho_i} =0 $, we obtain two roots of $\rho_i$, that is,
\begin{equation}\label{eq:tworoots}
	\rho^{\textmd{o}1} = \frac{1-\sqrt{1-\epsilon_i^2}}{\epsilon_i}\quad , \quad \rho^{\textmd{o}2}= \frac{1+\sqrt{1-\epsilon_i^2}}{\epsilon_i}.
\end{equation}
Observe that, with $0<\epsilon_i\le 1$, the first root satisfies $0 < \rho^{\textmd{o}1} \le 1$ while for the second root $\rho^{\textmd{o}2} \ge 1$. The equality signs of both roots hold, i.e., $\rho^{\textmd{o}1}=\rho^{\textmd{o}2}=1$, if and only if $\epsilon_i = 1$, therefore only the first root $\rho^{\textmd{o}1}$ is valid. This indicates that $g(\rho_i)$ is a convex function of the degree of impropriety, $\rho_i$, whose global minimum exists at $\rho_i = \rho^{\textmd{o}1}$. Upon summarizing the results for the two cases, this proves that the lower bound of the SNR gain is achieved if and only if \eqref{eq:cond} is satisfied.\hfill\qedsymbol

\textit{Remark 3}: Observe from \eqref{eq:SNRdelta3} that the SNR gain $\text{SNR}_{\Delta}^{\text{ap}}$ is a weighted sum of monotonically increasing functions (\textit{Case 1}) or convex functions (\textit{Case 2}), or their mixture, depending on the value of the \textit{power difference coefficient} $\epsilon_i$. This is in a very good agreement with {\textit{Theorem 2}} for small and medium degrees of impropriety $\rho_i$, as the SNR gain approximated by AUT, $\text{SNR}_{\Delta}^{\text{ap}}$ in \eqref{eq:SNRdelta1}, accurately describes the exact SNR gain, $\text{SNR}_{\Delta}$ in \eqref{eq:delta_snr}, as confirmed in the next section. 

{\textit{Remark 4}: For maximally improper noise with the degree of impropriety $\rho_i=1$, the function $g(\rho_i)$ in \eqref{eq:grhoi}, as well as the SNR gain $\text{SNR}_{\Delta}^{\text{ap}}$ in \eqref{eq:SNRdelta3}, approach infinity. This aligns with \textit{Remark 3} and suggests that an arbitrarily large SNR gain of WLMF under highly improper noise.


\section{Simulation of $\text{SNR}$ advantage of $\text{WLMF}$}
\label{sec:sim}
To demonstrate the proposed $\text{WLMF}$ concept and validate the SNR gain $\text{SNR}_{\Delta}$ and its lower bound, a simulation was conducted, whereby the noise $v(n)$ assumed the form of a moving average MA(2) process, given by
\begin{align}\label{eq:noise_model}
	v(n)= 0.9 u(n) - j0.1 u(n-1), 
\end{align}
where $u(n)$ was a zero-mean doubly white improper Gaussian process with variance $\sigma_u^2 = E[u(n)u^*(n)] = 1$ and complementary variance ${\tilde \sigma}_u^2 = E[u(n)u(n)] = \rho_u$, which gives a degree of impropriety\footnote{
Note that for an MA model with fixed taps as in \eqref{eq:noise_model}, the noise impropriety vector ${\bm \rho}$ is controlled by $\rho_u$, as all entries in ${\bm \rho}$ increase linearly against $\rho_u$.} $\rho_u$. 
\begin{figure}[t!]
	\centering
	\includegraphics[width=0.47\textwidth]{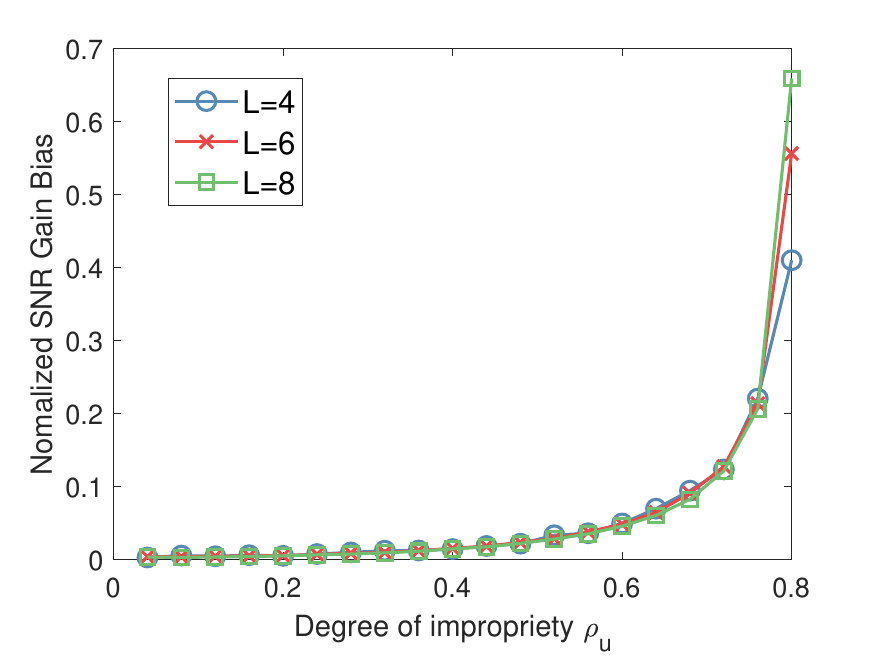}
	\caption{Normalized bias in SNR gain of WLMF over SLMF against the degree of impropriety $\rho_u$, which varies from an almost proper 0.04 to very improper 0.8.}
	\label{fig:gainbiasvsrho} 
\end{figure}

\noindent{\bf {SNR gain.}} The effectiveness of the approximated SNR gain $\text{SNR}_{\Delta}^{\text{ap}}$ in \eqref{eq:SNRdelta1} attained via AUT, was examined by using a zero-mean circular Gaussian process to generate a deterministic sequence $x(n)$ of length $N=100000$. The degree of impropriety $\rho_u$ was varying from an almost proper $\rho_u=0.04$ to very improper $\rho_u=0.8$. The length of the filters took the values $L\in\{4,6,8\}$. The difference between the approximated $\text{SNR}_{\Delta}^{\text{ap}}$ and the exact $\text{SNR}_{\Delta}$, that is, the bias in SNR, was measured by the normalized SNR bias, defined as
\[\frac{1}{N-L+1}\sum\limits_{n_p=L}^{N} \frac{\text{SNR}_{\Delta}^{\text{ap}}(n_p)-\text{SNR}_{\Delta}(n_p)}{\text{SNR}_{\Delta}(n_p)}.\]
Fig. \ref{fig:gainbiasvsrho} compares the so defined SNR bias against different degrees of impropriety $\rho_u$ and filter lengths $L$. Observe that the approximated SNR gain was statistically slightly larger than the exact SNR gain, but was a good match for the exact one for low and medium degrees of impropriety. This not only justifies the accuracy of the derived SNR gain expression in \eqref{eq:SNRdelta1}, but also verifies the lower bound derived in \textit{Theorem 2}, as the lower bound is usually achieved at a low degree of impropriety. In addition, this also shows that the filter length $L$ has a negligible impact on the approximation, except in the case of a very high degree of impropriety, where a larger filter length increases the approximation bias.
%
\begin{figure}[!t]
	\centering
	\includegraphics[width=0.47\textwidth]{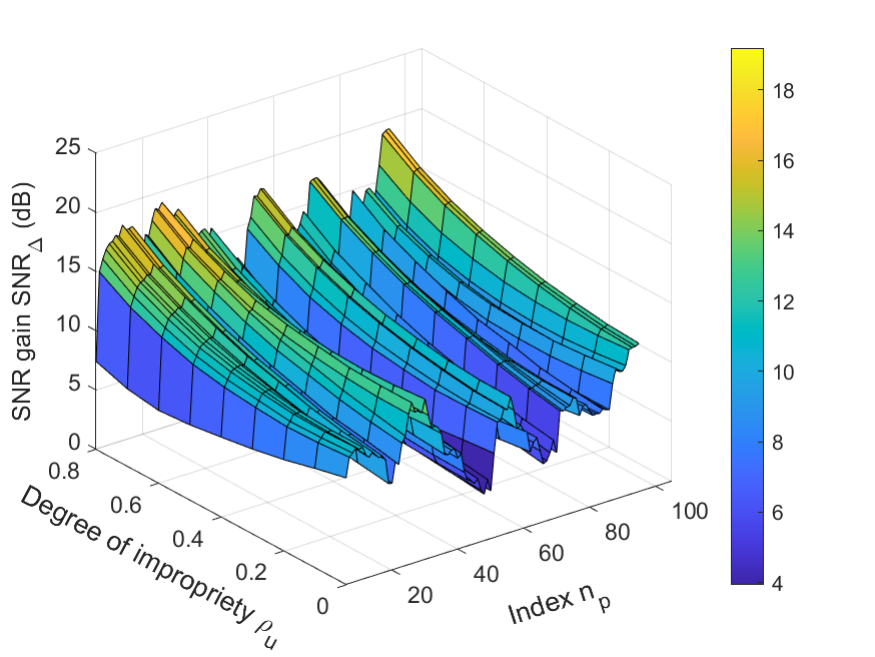}
	\caption{The SNR gain of an observed signal $r(n)$ generated by a deterministic input $x(n)$ of length $N=106$, corrupted by noise $v(n)$, of which the impropriety $\rho_u$ varies from an almost proper 0.04 to very improper 0.8.}
	\label{Fig:3D}
\end{figure}

\noindent{\bf {Lower bound.}} To verify the validity of the above derived lower bound, the filter length was set to $L=6$, and the degree of impropriety to $\rho_u = 0.5$. With these two parameters, the MA(2) process in \eqref{eq:noise_model} explicitly describes the noise covariance and complementary covariance matrices, ${\bf R}_v$ and ${\bf C}_v$, which with the help of AUT, give the diagonal matrices 
 %
\begin{align}
	 	{\bm \Lambda}_r &= \text{diag} \{0.98,0.93,0.86,0.77,0.70,0.65\},\\
	 	{\bm \Lambda}_c &= \text{diag} \{0.40,0.40,0.40,0.40,0.40,0.40\},
	\end{align}
and the corresponding impropriety vector 
\begin{equation}\label{eq:matchedrho}
	 {\bm \rho}=[0.41,0.43,0.46,0.51,0.56,0.60]^T.
\end{equation}
Then, according to \textit{Theorem 2}, we designed a $6 \times 1$ \textit{matched sequence} ${\bf{x}}_{\textmd {o}}=[0.77+j0.13,0.71+j0.25,-0.91-j0.33,-0.87-j0.07,-1.65-j0.62,0.74+j0.27]^T $, so that the \textit{power difference coefficients} $\epsilon_i$ of its rotated version, ${\bf{Q}}{\bf{x}}_{\textmd {o}}$, satisfy the condition in \eqref{eq:cond}, thereby achieving the lower bound of SNR gain at $\rho_u = 0.5$. The so designed \textit{matched sequence} ${\bf{x}}_{\textmd {o}}$ was concatenated with a $N=100$ long zero-mean circular Gaussian sequence, so as to form a $N=106$ long deterministic input $x(n)$. We generated the observed signal $r(n)$ with such an input $x(n)$ corrupted by noise $v(n)$, the impropriety parameter $\rho_u$ of which varied from 0.04 to 0.8. The signal $r(n)$ was fed into the SLMF and WLMF, respectively. Fig. \ref{Fig:3D} depicts the SNR gain, $\text{SNR}_{\Delta}$, which was computed by \eqref{eq:SNRGainDef} and averaged over $1000$ independent Monte Carlo trials. Observe the permanent existence of SNR gain advantage of WLMF over SLMF. Within the slice designated by $n_p=6$, where the input was the exact \textit{matched sequence} ${\bf{x}}_{\textmd {o}}$, it achieved its minimum at $\rho_u=0.5$. Moreover, Fig. \ref{Fig:3D} exhibits a mixture of monotonic and convex mappings between $\text{SNR}_{\Delta}$ and the degree of impropriety $\rho_u$. This demonstrates the validity of \textit{Theorem 2} and \textit{Remark 3}, while in each slice designated by index $n_p$, $\text{SNR}_{\Delta}$ becomes very large for highly improper noise, which conforms with \textit{Remark 4}.

\vspace{-1mm}
\section{Interpretation of complex-valued CNNs from the perspective of SLMF and WLMF}\label{sec:interprt}
\vspace{-1mm}
\begin{figure*}[!t]
	\centering
	\subfloat[]{ \label{Fig.MF_Input_signal}
		\includegraphics[width=0.33\textwidth]{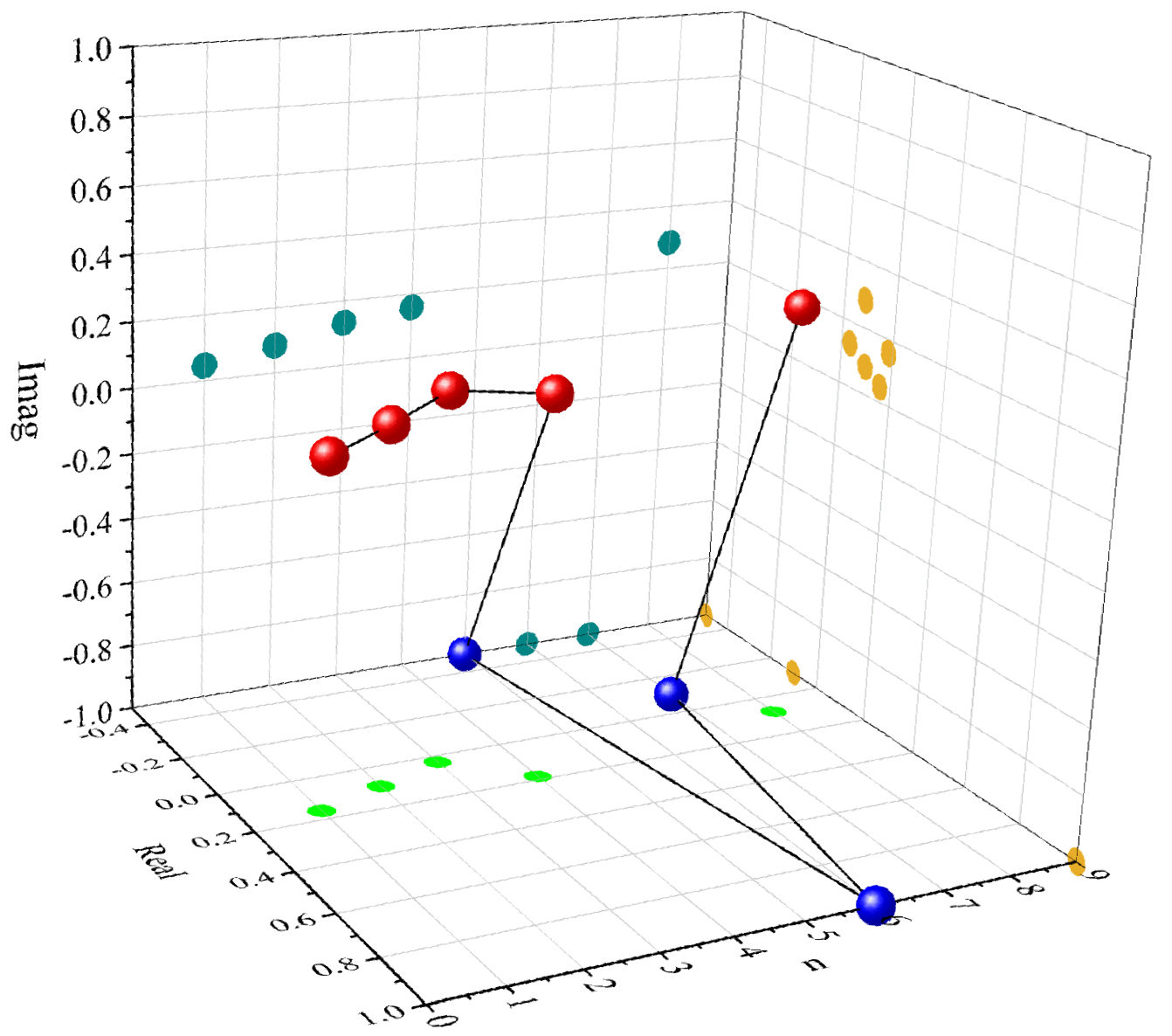}}
	\subfloat[]{\label{Fig.MF_pattern} 
		\includegraphics[width=0.33\textwidth]{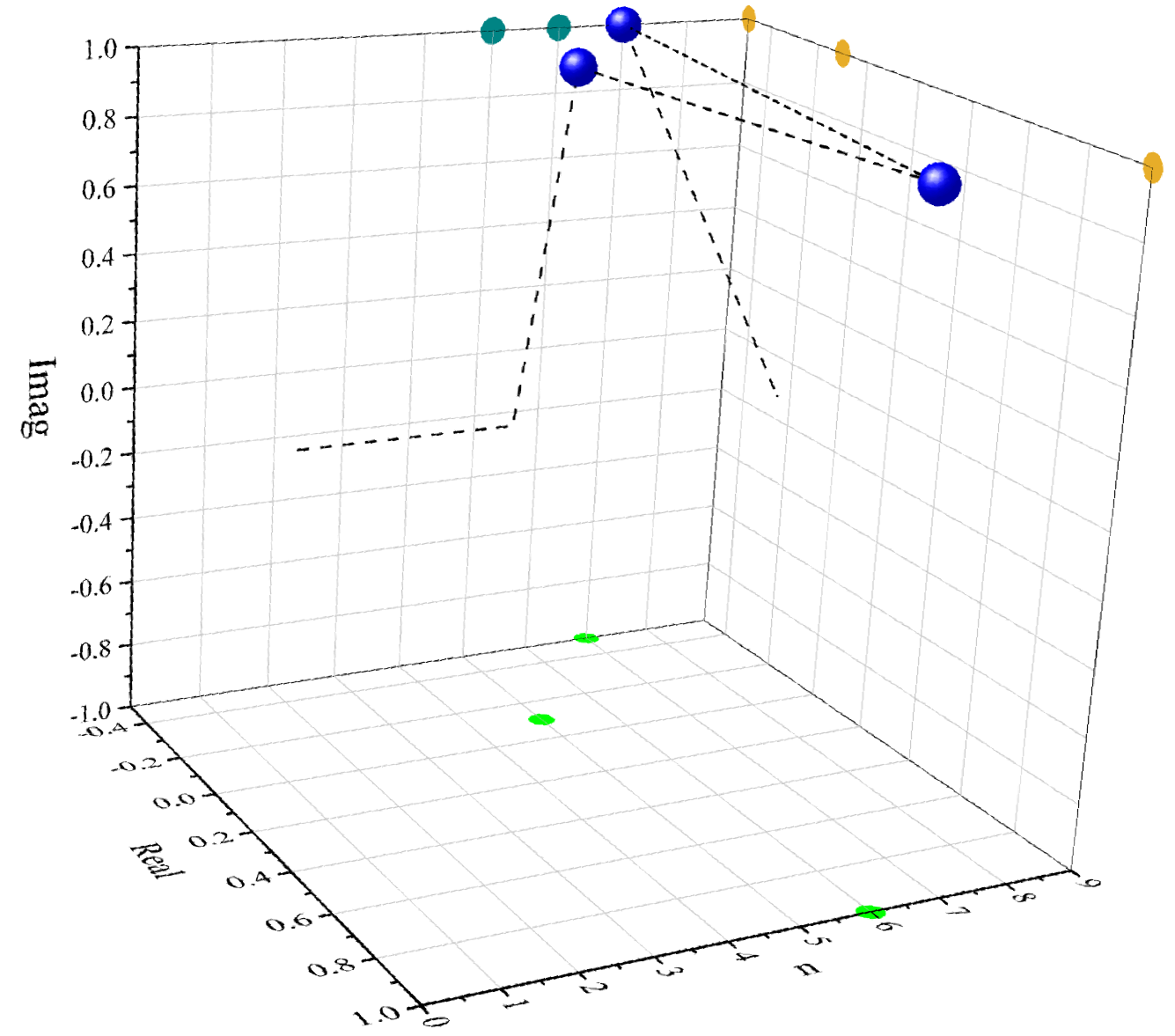}}
	\subfloat[]{\label{Fig.SL&WL_filter_output}
		\includegraphics[width=0.33\textwidth]{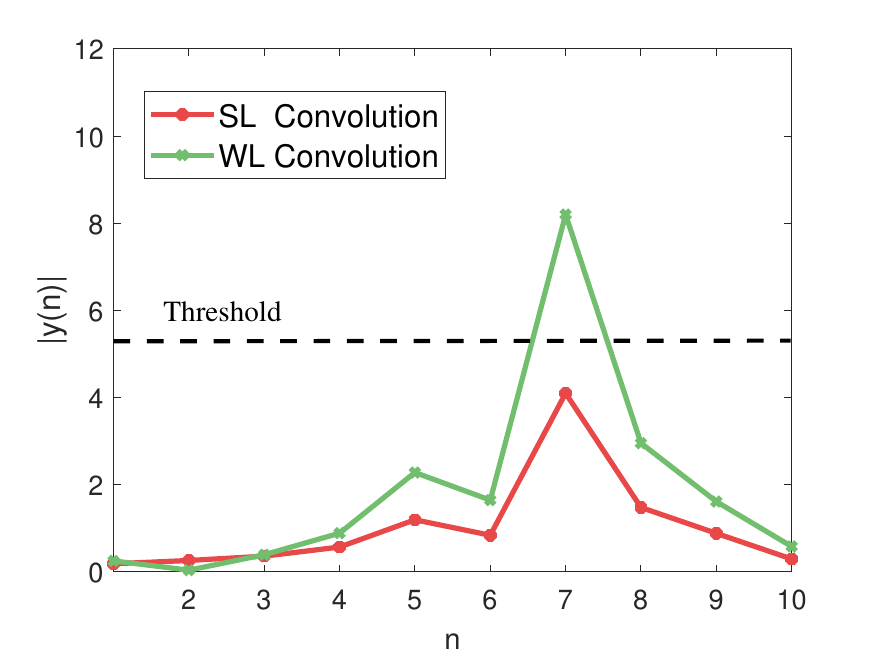}}
	\caption{Interpretation of complex-valued convolution as a matched filtering operation. (a) A noisy complex-valued input signal which contains a feature to be detected; the red dots represent noise and the blue dots the feature of interest corrupted by noise. (b) The MF template corresponding to the feature of interest, depicted in blue dots. (c) The outputs of WL and SL convolutional layers, with the maxima corresponding to the position of the identified feature of interest.}
	\label{fig:mfintp}
\end{figure*}

This section establishes an interpretation of the whole convolution-activation-pooling chain in complex-valued CNNs as a matched filtering operation. Building upon this insight, we subsequently demonstrate how the above proven performance advantage of WLMF over SLMF translates to enhanced classification capability of CNNs. Without loss of generality, this is shown for a simple complex-valued CNN consisting of a convolutional layer and a fully connected output layer. 

Let ${\bf{x}}^*_c$ denote the feature of interest in a complex-valued CNN, which is obtained by time-reversing and conjugating the MF template ${\bf{x}}$ \cite{Stankovic_Mandic_2023CNN}, namely ${\bf{x}}^*_c \triangleq [x^*(n_p-L+1), x^*(n_p-L+2),\dots,x^*(n_p-1),x^*(n_p)]^T$. Therefore, the SLMF operation in \eqref{eq:yn_p_SLMF} can be equivalently rewritten in the form of strictly linear convolution (SLC), given by
\begin{align}\label{eq:yn_p_SLConv}
	y(n_p)&={\bf{g}}^H{\bf{x}}^*_c+{\bf{g}}^H{\bf{v}},
\end{align}
where ${\bf{g}}\in {\mathbb{C}}^{L\times 1}$ denotes an $L$-tap convolutional filter, which, from the perspective of SLMF, equals the ``matched filtering'' coefficients $\hat {\bf{g}}=\alpha {\bf{R}}_{v}^{-1}{\bf{x}}^*_c$, when the filter output $y(n_p)$ achieves the maximum SNR. Analogously, the WLMF operation in \eqref{eq:yn3} can be transformed to widely linear convolution (WLC) as
\begin{equation}\label{eq:yn_pWLConv}	
	y(n_p)={\bf{g}}_1^H{\bf{x}}^*_c+{\bf{g}}_2^H{\bf{x}}_c+{\bf{g}}_1^H{\bf{v}}+{\bf{g}}_2^H{\bf{v}}^*.
\end{equation}

\textit{Remark 5}: The above transformation reveals that either the SLMF in \eqref{eq:yn_p_SLMF} or WLMF in \eqref{eq:yn3}, which maximize the cross-correlation between the observed signal $r(n)$ and the input $x(n)$, is equivalent to their corresponding convolution-based implementation, given by \eqref{eq:yn_p_SLConv} or \eqref{eq:yn_pWLConv}. This equivalence illuminates the underlying principle that the convolution of a signal with a target feature, the very essence of the convolutional layer in CNNs, acts as a matched filtering mechanism which detects the existence and location of that characteristic feature within the data. It is also worth noting that our interpretation in the complex-valued context is different from its real-valued counterpart introduced in \cite[Sec. II]{Stankovic_Mandic_2023CNN}, which requires only time-reversing the template ${\bf{x}}$. 

Fig. \ref{fig:mfintp} visualizes the interpretation from \textit{Remark 5}. A noisy complex-valued input signal $x(n)$ of length $N = 8$ is shown in Fig. \ref{Fig.MF_Input_signal} as a solid line, whereby the red points contain only noise and the blue points represent a feature of length $L = 3$, ${\bf{x}}^*_c = [-0.5-j1, 1-j1, -0.1-j1]$, corrupted by noise. The MF template corresponding to the feature of interest, ${\bf{x}} = [-0.1+j1, 1+j1, -0.5+j1]$, implemented by the MF operations as in \eqref{eq:yn_p_SLMF} and \eqref{eq:yn3}, is shown in Fig. \ref{Fig.MF_pattern}. We utilized both SLC and WLC to detect the presence of the feature in Fig. \ref{Fig.MF_pattern}, and both of their outputs, quantified by the modulus, are given in Fig. \ref{Fig.SL&WL_filter_output}, along with an appropriate threshold in dashed line. As desired, both SLMF and WLMF operations achieved their maxima at the matching location $n=7$, whereby the WLC exhibited a larger peak than the SLC. Given our interpretation of CNNs as performing matched filtering operations, this difference reflects the proven advantage of WLMF over SLMF.

\textit{Remark 6}: Observe that the only important sample in the output of the SLMF or WLMF in Fig. \ref{Fig.SL&WL_filter_output} is the maximum one, whose value and position are retained for further processing (a following layer of a CNN). Also notice that its value and position remain invariant upon applying a rectifying function, such as the rectified linear unit (ReLU), or after maximum pooling which selects the maximum sample from a sequence. In this way the SLMF and WLMF provide physical interpretation of the whole convolution-activation-pooling chain in complex-valued CNNs.
\begin{figure}[!t]
	\centering
	\includegraphics[width=0.48\textwidth]{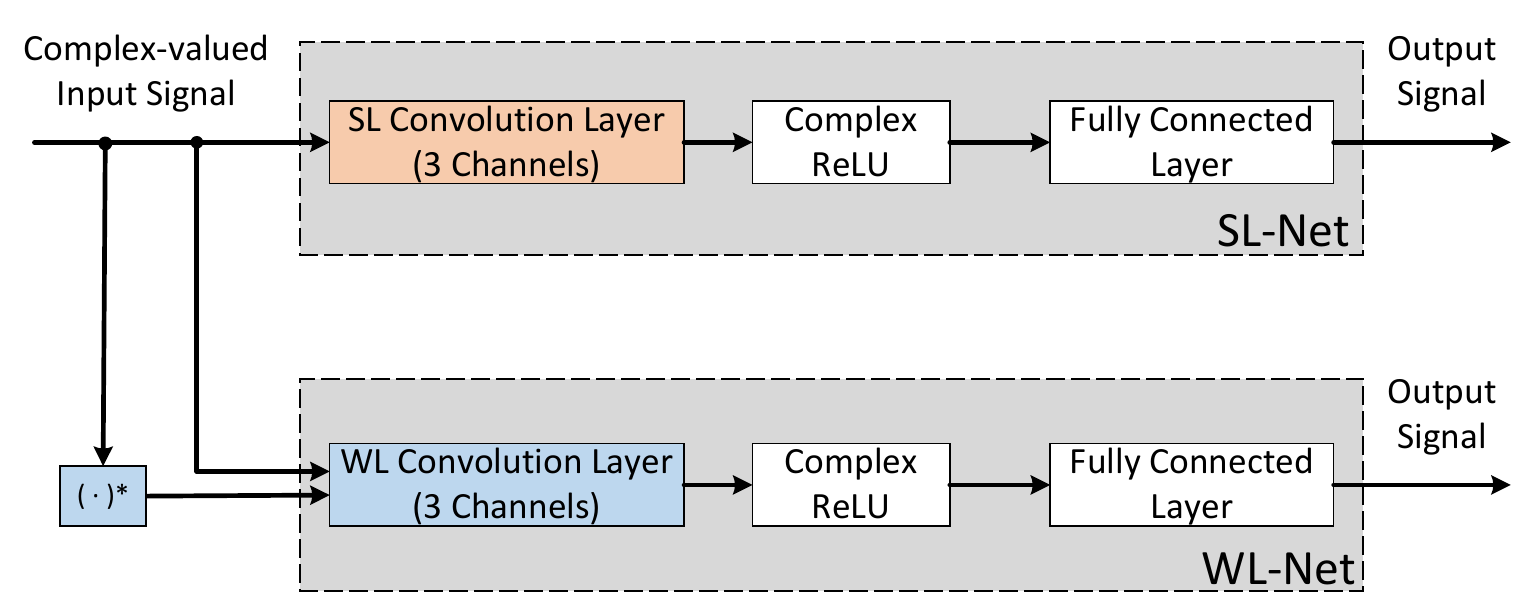}
	\caption{Block diagram of two simple complex-valued CNNs used in the simulations. The SL-Net employs standard complex input and SLC, and while the WL-Net uses augmented complex input and WLC.}
	\label{Fig:CNN_diagram}
\end{figure}
%
Since according to \textit{Remark 6}, the larger maximum of the WLMF over SLMF in Fig. \ref{Fig.SL&WL_filter_output} indicates a better feature detection ability of WLMF, we further validated the advantage of WLC over SLC in a typical identification task performed by complex-valued CNNs. As depicted in Fig. \ref{Fig:CNN_diagram}, two simple examples of such CNNs were considered: one employing standard input and SLC, named the SL-Net, and the other using an augmented input and WLC, referred to as WL-Net. Both networks operated on the task of identifying the presence of two complex-valued temporal patterns:
\begin{itemize}
	\item ${\text{Pattern 1}} = [-0.5-j1, 1-j1, -0.5-j1] + {\bf{v}}$, for which the target output was $
	{\bf{t}} = [1,0]^T$;
	\item ${\text{Pattern 2}} = [1+j1, 1+j1, 1+j1]+{\bf{v}}$, for which the target output was ${\bf {t}} = [0,1]^T$,
\end{itemize}
where the complex-valued noise vector ${\bf{v}}={\bf{v}}_r+j{\bf{v}}_i$, and, for both networks, each entry of the real-valued vectors, ${\bf{v}}_r$ and ${\bf{v}}_i$, was drawn from an independent identically distributed uniform distribution with values in the region $[0,0.3]$. The length of the input $x(n)$ was set to $N = 8$, and these noisy signals were further corrupted by doubly white circular Gaussian noise with a standard deviation of $0.05$, and then normalized to unit energy. At the convolutional layer of each CNN, convolution filters of 3 samples length were used in each of the 3 convolutional channels. In the nonlinear activation phase, the overall complex-valued output for every channel of the convolutional layer was split into its respective real and imaginary components, each with a real-valued bias term, before a ReLU activation function was applied, i.e., $f(z) = {\text{ReLU}}(\Re\{z\}+b_r) + j {\text{ReLU}}(\Im\{z\}+b_i)$. The SoftMax activation function was used at the fully connected layer, with two output neurons that correspond to the two considered patterns in the target signal ${\bf{t}}$, as described above. 

\begin{figure}[!t]
	\centering
	\includegraphics[width=0.47\textwidth]{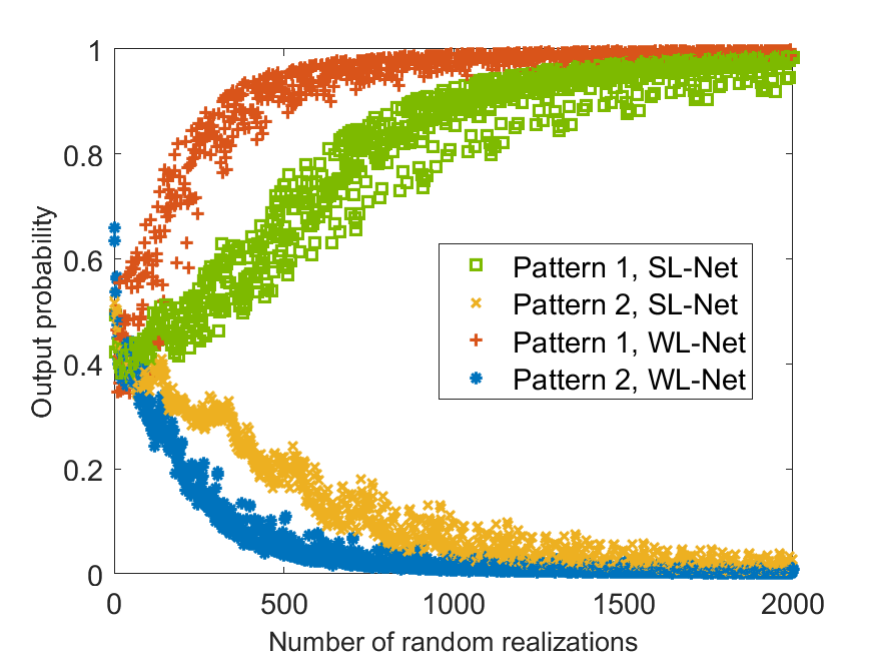}
	\caption{Output probabilities of the two considered  CNNs, SL-Net and WL-Net, which operate on the task of identifying the presence of two patterns.}
	\label{Fig:CNN}
\end{figure}
The two networks in Fig. \ref{Fig:CNN_diagram} were trained through backpropagation on a computer equipped with an Intel Core CPU i7-8550U @ 1.80GHz, and the training was performed over 10 epochs of 200 random signal realizations on Pytorch 1.13.1 and Python 3.8.2 platforms. The output probabilities of the SL-Net and WL-Net are depicted in Fig. \ref{Fig:CNN}. Observe that the WL-Net achieved perfect classification considerably earlier than the SL-Net in identifying both Pattern 1 and Pattern 2. The times required for the execution by the SL-Net and WL-Net were respectively 16.32 sec and 15.88 sec. Given that, as revealed in \textit{Remark 5}, the convolutional layer in CNNs performs precisely the matched filtering operation, despite twice the number of parameters over SL-Net, the reduced training time offered by the WL-Net, shown in Fig. \ref{Fig:CNN}, can be explained by the SNR advantage of WLMF over SLMF proven in our analysis in Sec. \ref{sec:ana}. Indeed, since the two complex-valued patterns in the identification task were corrupted by circular noise, Fig. \ref{Fig:CNN} is in agreement with \textit{Corollary 1}, which gives the exact SNR gain of WLMF over SLMF in the case of proper complex noise. This numerical result serves to comprehensively and systematically demystify the operation of complex-valued CNNs from the perspective of WLMF. 

\vspace{-1mm}
\section{Conclusion}
\vspace{-1mm}
Real-valued CNNs have revolutionized the way we approach image processing applications, however, despite their enormous potential, similar results for complex-valued CNNs are few and far between. This is particularly the case for time series data, which are notoriously difficult to process using CNNs due to their nonstationarity, presence of noise and the lack of physical interpretability of the solution. To address these issues, we have first introduced a general WLMF model for deterministic inputs and have provided its solution by maximizing the output SNR. The SNR gain of the WLMF over SLMF has been rigorously analyzed, which demonstrates that the WLMF always enhances the SNR over SLMF, except for a null input. Next, a lower bound on the SNR gain of WLMF has been established, and the condition to achieve this bound has been explicitly given in a closed form. This has been enabled physical interpretability of the convolution-activation-pooling chain within CNNs based on the first principle of pattern matching through complex matched filters---the methodology routinely used in radar, sonar, communications and biomedical engineering. Simulations in both widely linear complex matched filtering setting and for complex-valued CNNs support the analysis. It is our hope that this result will pave the way for the design and analysis, of otherwise black box CNNs, starting from first principles, and in this way empower the research communities working on CNNs with a much needed interpretability and design framework.


\bibliographystyle{IEEEtran}
\bibliography{reference_abbr}
\end{document}